\documentclass{article}
\usepackage{subfigure}
\usepackage{multirow}
\usepackage{spconf,amsmath,graphicx,amssymb,amsthm,dsfont,amsfonts,bm}
\usepackage[acronym]{glossaries}
\usepackage{caption}
\usepackage{amsmath}
\newacronym{knn}{KNN}{K nearest neighbors}
\newacronym{glr}{GLR}{graph Laplacian regularization}
\newacronym{cd-sgw}{CD-SGW}{spectral graph wavelet-based color denoising}
\newacronym{3dpbs}{3DPBS}{3-dimensional patch-based similarity}
\newacronym{lbvh}{LBVH}{linear bounding volume hierarchy}
\newacronym{fgbd}{FGBD}{fast graph-based denoising}
\newacronym{slg}{SLG}{scan-line graph}
\newacronym{ne-gbp}{NE-GBP}{noise estimation using graph-based patches}
\newacronym{fslr}{FSLR}{filter selection with limited region}
\newacronym{pca}{PCA}{principal component analysis}
\newacronym{mvub}{MVUB}{Microsoft voxelized upper bodies}
\newacronym{psnr}{PSNR}{Peak signal-to-noise ratio}
\newacronym{bf-knn}{BF-KNN}{Brute-force KNN}
\newacronym{megw}{MEGW}{median estimator with graph wavelets}

\title{FAST GRAPH-BASED DENOISING FOR POINT CLOUD COLOR INFORMATION}
\name{
{
Ryosuke Watanabe $^{\dagger,\ddagger}$, 
Keisuke Nonaka $^{\ddagger}$,
Eduardo Pavez $^{\dagger}$,
Tatsuya Kobayashi $^{\ddagger}$, 
Antonio Ortega $^{\dagger}$
}
}
\address{
$^{\dagger}$ University of Southern California,
$^{\ddagger}$ KDDI Research, Inc.  
}

\begin{document}

\maketitle
\begin{abstract}
Point clouds are utilized in various 3D applications such as cross-reality (XR) and realistic 3D displays.
In some applications, e.g., for live streaming using a 3D point cloud, real-time point cloud denoising methods are required to enhance the visual quality. 
However, conventional high-precision denoising methods cannot be executed in real time for large-scale point clouds owing to the complexity of graph constructions with K nearest neighbors and noise level estimation.
This paper proposes a fast graph-based denoising (FGBD) for a large-scale point cloud.
First, high-speed graph construction is achieved by scanning a point cloud in various directions and searching adjacent neighborhoods on the scanning lines.
Second, we propose a fast noise level estimation method using eigenvalues of the covariance matrix on a graph.
Finally, we also propose a new low-cost filter selection method to enhance denoising accuracy to compensate for the degradation caused by the acceleration algorithms.
In our experiments, we succeeded in reducing the processing time dramatically while maintaining accuracy relative to conventional denoising methods. Denoising was performed at 30fps, with frames containing approximately 1 million points.
\end{abstract}

%
%
%
%
%
%%%%%%%%%%%%%%%%%%%%%%%%%%%%%%%%%%%%%%%%%%%%%%%%%%%%%%%%%%%%%%%%%%%%%%%
%                           1. INTRODUCTION
%%%%%%%%%%%%%%%%%%%%%%%%%%%%%%%%%%%%%%%%%%%%%%%%%%%%%%%%%%%%%%%%%%%%%%%

\renewcommand{\thefootnote}{\fnsymbol{footnote}}

\footnote[0]{
  Copyright~\copyright~2024 IEEE. Personal use of this material is permitted. Permission from IEEE must be obtained for all other uses, in any current or future media, including reprinting/republishing this material for advertising or promotional purposes, creating new collective works, for resale or redistribution to servers or lists, or reuse of any copyrighted component of this work in other work.}

\section{Introduction}
\label{sec:Introduction}
Point clouds are utilized in a variety of 3D applications such as cross-reality (XR)~\cite {telepresence} and holographic 3D displays~\cite {Holo3DTV}.
In these applications, scanned point clouds, consisting of a collection of 3D coordinates and associated color signals,  are often perturbed by noise caused by sensor measurement errors.
Thus, point cloud denoising methods are important to improve the accuracy of downstream tasks such as object detection~\cite{2018PIXOR}, action recognition~\cite{ActionRecog2022}, and point cloud compression~\cite{PCC2018}.
In many applications, denoising must be executed in real time.
For example, in a point cloud streaming scenario for a 3D telepresence system, denoising is required just after scanning to reduce noise caused by a sensor, and after receiving data on the user side to suppress noise caused by compression errors.
If a point cloud is scanned and transmitted in real time for 3D telepresence, denoising must be executed in real time.

Some methods have been proposed to reduce noise on point cloud geometry~\cite{GPDNet, DMRDenoise, IBR, GBPCD, GeomDenoising}, while others focus on color denoising ~\cite{MS-GAT, DineshGLRGTV2019, GAC, 3DPBS, PCS2022}.
Since both geometry and color information directly affect visual quality, they are important to enhance the user experience in 3D applications.  
%In principle, 
Though our proposal can be applied to both color and geometry denoising, here we focus on point cloud color denoising and leave geometry denoising for future work.
Deep learning-based and graph-based approaches have been among the most widely studied techniques for point cloud color denoising in recent years. 
We focus on graph-based methods~\cite{DineshGLRGTV2019, GAC, 3DPBS} because, unlike deep learning-based methods~\cite{MS-GAT}, training data are not required.
Recent graph-based approaches include \gls{glr}~\cite{DineshGLRGTV2019}, \gls{cd-sgw}~\cite{GAC}, and \gls{3dpbs}~\cite{3DPBS}.
\gls{glr} \cite{DineshGLRGTV2019} utilizes graph Laplacian regularization as a smoothness prior to achieve accurate color denoising. 
\gls{cd-sgw}~\cite{GAC} utilizes BayesShrink \cite{BayesShrinkImage}, a popular wavelet shrinking technique for image denoising, to reduce the high-frequency wavelet components in the graph spectral domain.
In \gls{3dpbs}~\cite{3DPBS}, to improve the denoising accuracy, a graph construction method that is not susceptible to noise has been proposed.
However, since detailed 3D models in these applications often result in large-scale point clouds with hundreds of thousands of points  \cite{8iVFBv2,MVUB}, conventional denoising methods \cite{DineshGLRGTV2019,GAC,3DPBS} cannot be performed in real time.
This is due primarily to the large computation times required by \gls{knn} graph constructions in graph-based methods \cite{DineshGLRGTV2019,GAC,3DPBS}.

Many methods to speed up graph construction, in particular, \gls{knn}, have been proposed \cite{HighDimKNN1,HighDimKNN2,BN-KNN,CUDA-KNN,KDtreeNN,LBVH}.
Some of them  \cite{HighDimKNN1,HighDimKNN2} work well for high-dimensional data but are less effective for 3D data such as point clouds.
Although parallel computing methods with a GPU \cite{BN-KNN,CUDA-KNN} are proposed, the processing time is still large because they calculate distances between points by brute-force approach.
To construct a graph from a point cloud, a neighbor search from each point is required to decide the connectivity.
Fast methods build a data structure suitable for neighborhood search, e.g., kdtree~\cite{KDtreeNN} or \gls{lbvh}~\cite{LBVH}, in advance, speeding up the neighbor search process dramatically compared with the brute-force methods \cite{BN-KNN,CUDA-KNN}.
These methods are most efficient when the data structures do not change.
However, in point cloud video, a new data structure is needed for each frame, so these methods are not as advantageous.

\begin{figure}[t]
\centering 
   %%%%%%%%%%%%%%%%%%%%%%%%%%%%%%%%%%%%%%%%%%%%%%%%%%%%%%%%%%%%%%%%%%%%%
   \subfigure[]{
   \includegraphics[width=.90\columnwidth]{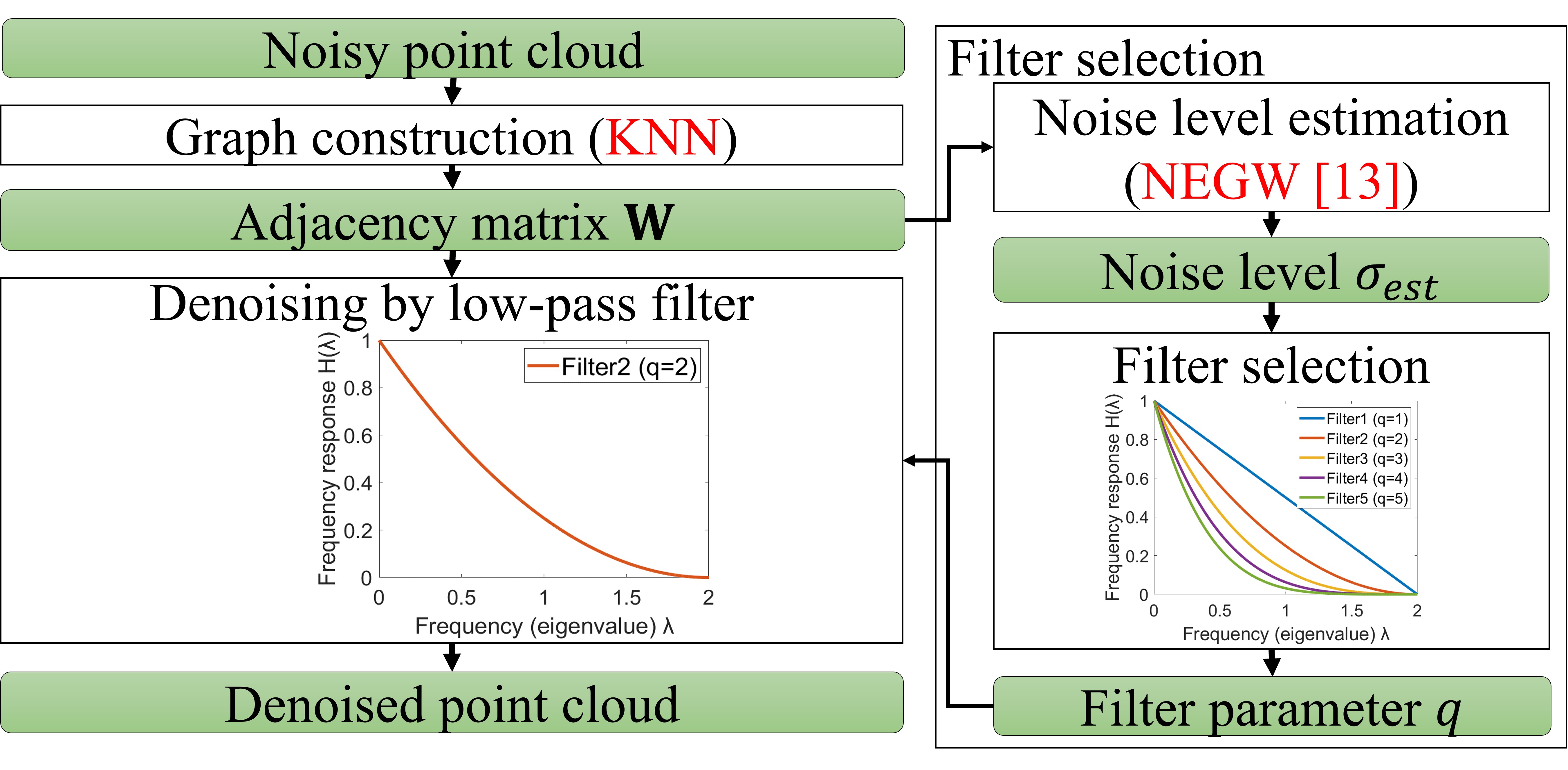}
   \label{fig:flow_3dpbs}
  }
  \subfigure[]{
   \includegraphics[width=.90\columnwidth]{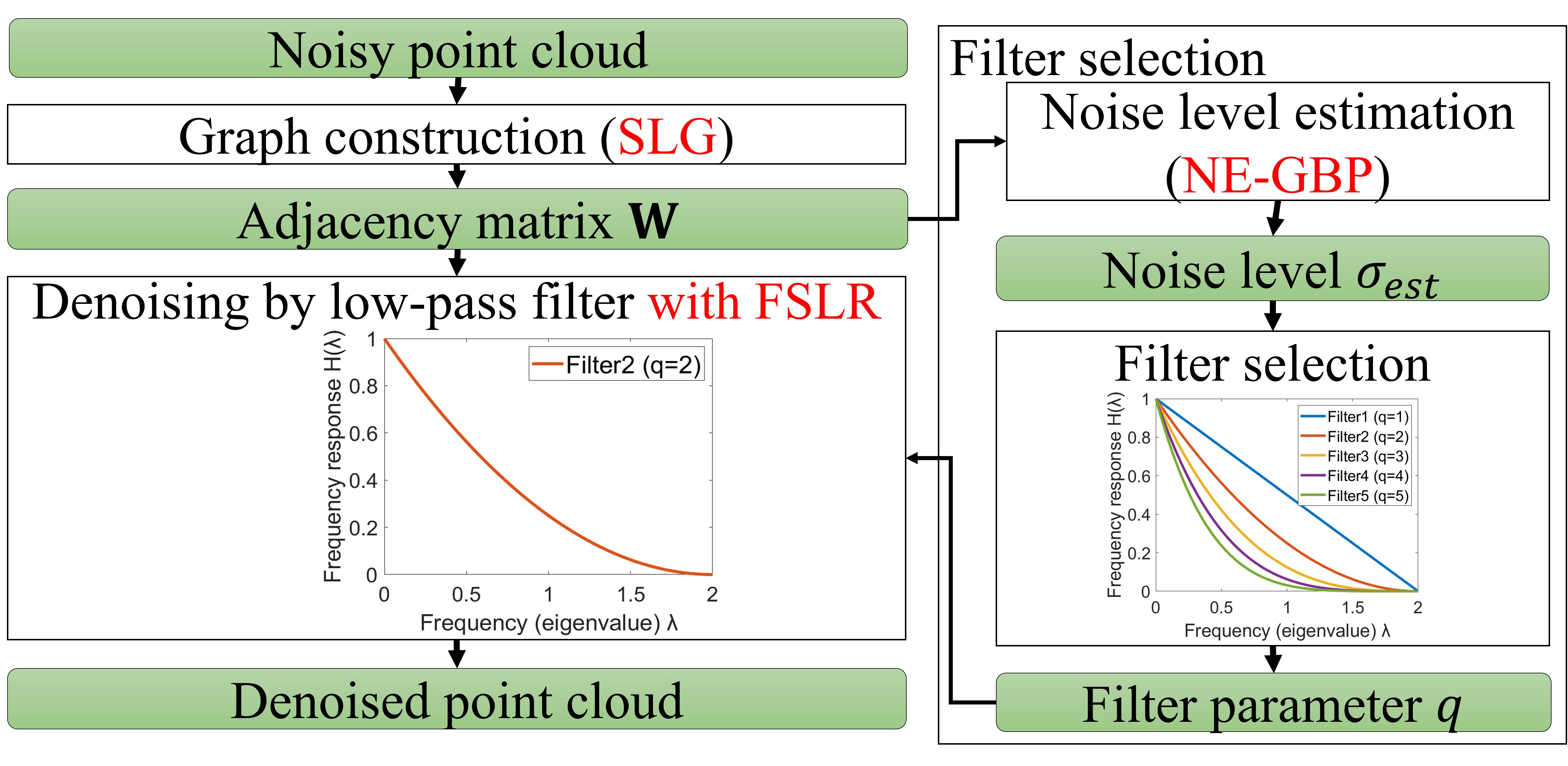}
   \label{fig:flow_fgbd}
  }
  %%%%%%%%%%%%%%%%%%%%%%%%%%%%%%%%%%%%%%%%%%%%%%%%%%%%%%%%%%%%%%%%%%%%%
  \caption{Basic flow of (a) 3DPBS~\cite{3DPBS} and (b) FGBD (proposed).}
\label{fig:flow}
\end{figure}

The proposed method called \gls{fgbd} realizes real-time denoising for a large-scale point cloud.
The starting point of \gls{fgbd} is our recent research called \gls{3dpbs}~\cite{3DPBS}, which is a high-precision and large-complexity color denoising method.
Fig.\ref{fig:flow} describes that \gls{3dpbs} is composed of three major processes: 1) \gls{knn}-based graph construction, 2) low-pass filter selection based on an estimated noise level, and 3) low-pass filter execution on graph spectral domain.
As discussed above, \gls{knn} graph construction has a large complexity. 
In addition, since the noise level is estimated by a conventional SGWT-based noise estimation method \cite{GAC}, the noise estimation process has significant complexity.
Although noise estimation may not be required in all frames if there is no change in the noise level, it frequently changes in practice, e.g., noise caused by the scanning sensor's heat. Thus, high-speed processing is desirable to respond quickly to the changes.
Finally, though \gls{3dpbs} utilizes polynomial approximation in the low-pass filter execution process to represent flexible frequency response, the approximation has some complexity.
The previous study \cite{PCS2022} can solve the complexity of the low-pass filter execution. However, denoising accuracy is degraded.
Our study, \gls{fgbd}, solves these three major problems regarding complexity without significant denoising performance degradation compared with conventional accurate denoising methods. 

\footnote[0]{This work was supported by the Ministry of Internal Affairs and Communications (MIC) of Japan (Grant no. JPJ000595).}
\renewcommand{\thefootnote}{\arabic{footnote}}

Our paper's contributions are as follows: 1) A \textbf{\gls{slg} construction} that does not require building a data structure in advance.
2) A fast and accurate \textbf{\gls{ne-gbp}} based on \gls{pca} on a set of neighborhoods.
3) An accurate \textbf{\gls{fslr}} which improves low-pass filter selection scheme proposed by \gls{3dpbs}~\cite{3DPBS} with low-cost processing. 
The choice of low-pass filter's frequency response is improved by not using some regions where the noise-free signals have high-frequency components (e.g., regions where a few sharp color changes).

While contributions 1) and 2) mainly focus on improving computation speed, contribution 3) is a proposal to improve denoising accuracy with
a slight increase in processing time.
In the experiments, \gls{fgbd} is carried out with a single GPU based on NVIDIA CUDA implementation. 
We report that the processing of \gls{fgbd} was 1500 times faster than that of \gls{3dpbs} with 8iVFBv2 \cite{8iVFBv2} and MVUB \cite{MVUB} datasets while maintaining the denoising accuracy.
As a result, denoising was performed in real time at 30fps with a point cloud composed of around 1 million points.
Note that our proposed methods, particularly SLG, can also be applied to accelerate geometry denoising using graph-based methods.
Furthermore, SLG has the potential to contribute to speeding up not only denoising but also various applications that utilize a graph.

%
%
%
%
%
%%%%%%%%%%%%%%%%%%%%%%%%%%%%%%%%%%%%%%%%%%%
%       2. PROPOSED METHOD
%%%%%%%%%%%%%%%%%%%%%%%%%%%%%%%%%%%%%%%%%%%
\section{PROPOSED DENOISING METHOD}
\label{sec:proposed-method}
\subsection{Overview of fast graph-based denoising (FGBD)}
\label{sec:proposed-method-overview}
\begin{figure}[t]
\centering 
   \includegraphics[width=0.90\linewidth]{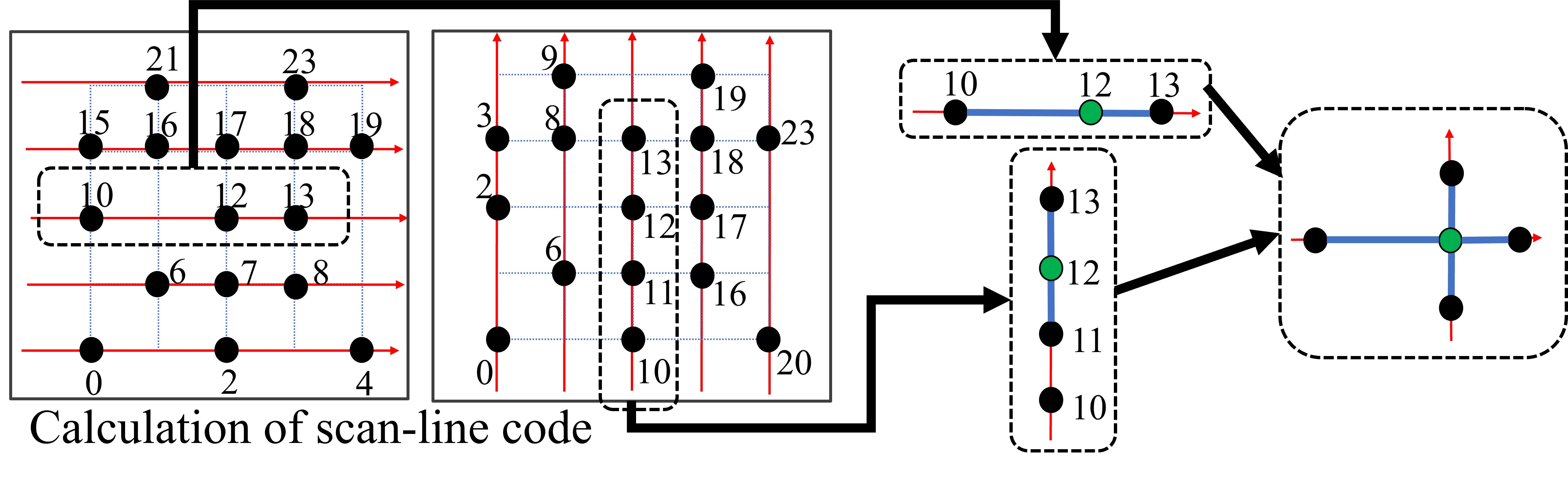}
   \caption{Neighbor search on scan-line codes (Example of 2-D).}
\label{fig:SLG2D}
\end{figure}

Fig.\ref{fig:flow} shows the overall calculation flow of the proposed method called \gls{fgbd}, which includes the following 4 steps.
%
%%%%%%%%%%%%%%%%%%%%%%%%%%%%%%%%%%%%%%%%%%%%%%%%%%%%%
%
\begin{enumerate}
\setlength{\itemsep}{0cm}
\setlength{\leftskip}{-0.4cm}
%%%%%%%%(1)%%%%%%%
\item 
A graph is constructed from a noisy point cloud by using a scan-line graph (SLG).
%%%%%%%%(2)%%%%%%%
\item 
Noise level $\sigma_{\mathrm{est}}$ is estimated by the method called noise estimation using graph-based patches (NE-GBP).
%%%%%%%%(3)%%%%%%%
\item 
The parameter of a graph low-pass filter is determined by using filter selection with limited region (FSLR).
%%%%%%%%(4)%%%%%%%
\item
Color attributes are denoised using a vertex domain low-pass graph filter \cite{PCS2022}.
\end{enumerate}
%
%%%%%%%%%%%%%%%%%%%%%%%%%%%%%%%%%%%%%%%%%%%%%%%%%%%%%
%
We explain steps 1 and 2 in Section \ref{sub-sec:proposed-scan-line-graph} and Section \ref{sec:proposed-NEGBP}, respectively.
Finally, steps 3 and 4 are described in Section \ref{sec:filter-selection}.
\subsection{Graph construction based on SLG}
\label{sub-sec:proposed-scan-line-graph}

We introduce a high-speed graph construction method called \gls{slg}.
A graph is constructed from a given noisy point cloud $P = \{p_{i}\}^N_{i=1}$, where point $p_i$ has coordinate $\bm{g}_i \in \mathbb{R}^{3}$ and attribute signals $\bm{f}_i \in \mathbb{R}^{3}$.
\gls{slg} assumes that the coordinate signals $\boldsymbol{g}_i$ are quantized to integers with $b$ bits.
Since most point clouds in real-time applications go through voxelization, they have integer coordinates \cite{PCC2018}.
If they are represented by floating points, the coordinate signals $\bm{g}_i$ are quantized to integers before starting the graph construction 

Fig.~\ref{fig:SLG2D} shows the graph construction process of \gls{slg}.
First, a one-dimensional code (called scan-line code) $C_{i,l}$, where $l$ indicates the index of a line, is calculated from $\bm{g}_i$.
For example, scan-line codes calculated by raster scan are expressed by
\begin{eqnarray}
\label{eq:SLG3-1}
C_{i,1} = 2^{2b} \boldsymbol{g}_{i,z} + 2^b \boldsymbol{g}_{i,y} + \boldsymbol{g}_{i,x}, \\
\label{eq:SLG3-2}
C_{i,2} = 2^{2b} \boldsymbol{g}_{i,x} + 2^b \boldsymbol{g}_{i,z} + \boldsymbol{g}_{i,y}, \\
\label{eq:SLG3-3}
C_{i,3} = 2^{2b} \boldsymbol{g}_{i,y} + 2^b \boldsymbol{g}_{i,x} + \boldsymbol{g}_{i,z},
\end{eqnarray}
where $\boldsymbol{g}_{i,x}$ shows the $x$-coordinate value of $\boldsymbol{g}_{i}$.
Second, points are ordered based on each scan-line code $C_{i,l}$ by a GPU-friendly sort algorithm, called radix sort \cite{radixsort}. Next, two consecutive points on the scan-line code are regarded as neighbors and connected in a graph.
This process is carried out independently for each code.
In other words, 6 neighborhoods (= 2 neighbors $\times$ 3 codes) are obtained when we introduce 3 codes represented by \eqref{eq:SLG3-1}, \eqref{eq:SLG3-2}, and \eqref{eq:SLG3-3}.
Finally, after acquiring the connectivity by SLG, the adjacency matrix $\mathbf{W}=\{w_{ij}\}$ is calculated by using the Gaussian kernel as follows.%
\begin{equation}
\label{eq:BF-weight}
w_{ij} = \exp \left(-\frac{(\boldsymbol{g}_i - \boldsymbol{g}_j)^2}{\sigma^{2}_{g}}\right).
\end{equation}
If $\sigma_{g}$ is very large (respectively small) relative to the numerator, all the weights become almost 1 (respectively 0). 
To avoid those two extreme cases, $\sigma_{g}$ is computed as the average Euclidean distance between $p_i$ and its neighbors.
Note that when a raster scan is performed, a distant connection may occur when a scan line connects to the next line.
In this case, the Euclidean distance becomes too large and the similarity is close to zero in $\mathbf{W}$. 
Thus, this connection does not cause an adverse effect.

%The proposed method can be generalized as the idea of converting coordinate signals into some one-dimensional codes and neighbors are detected on the codes. 
Since neighbors can be obtained by only sorting, computing connectivity in a graph no longer requires many distance calculations or building a specific data structure such as kdtree.
In our preliminary experiments, we tried other types of codes, e.g., Morton codes \cite{PCC2018} and diagonal scan-lines instead of \eqref{eq:SLG3-1}, \eqref{eq:SLG3-2}, and \eqref{eq:SLG3-3}.
However, the three line-scan codes introduced above provide the best trade-off between accurate denoising and construction speed. Therefore, $C_{i,1}, C_{i,2}$ and $C_{i,3}$ are adopted in this paper.

\begin{figure}[t]
\centering 
   \includegraphics[width=0.90\linewidth]{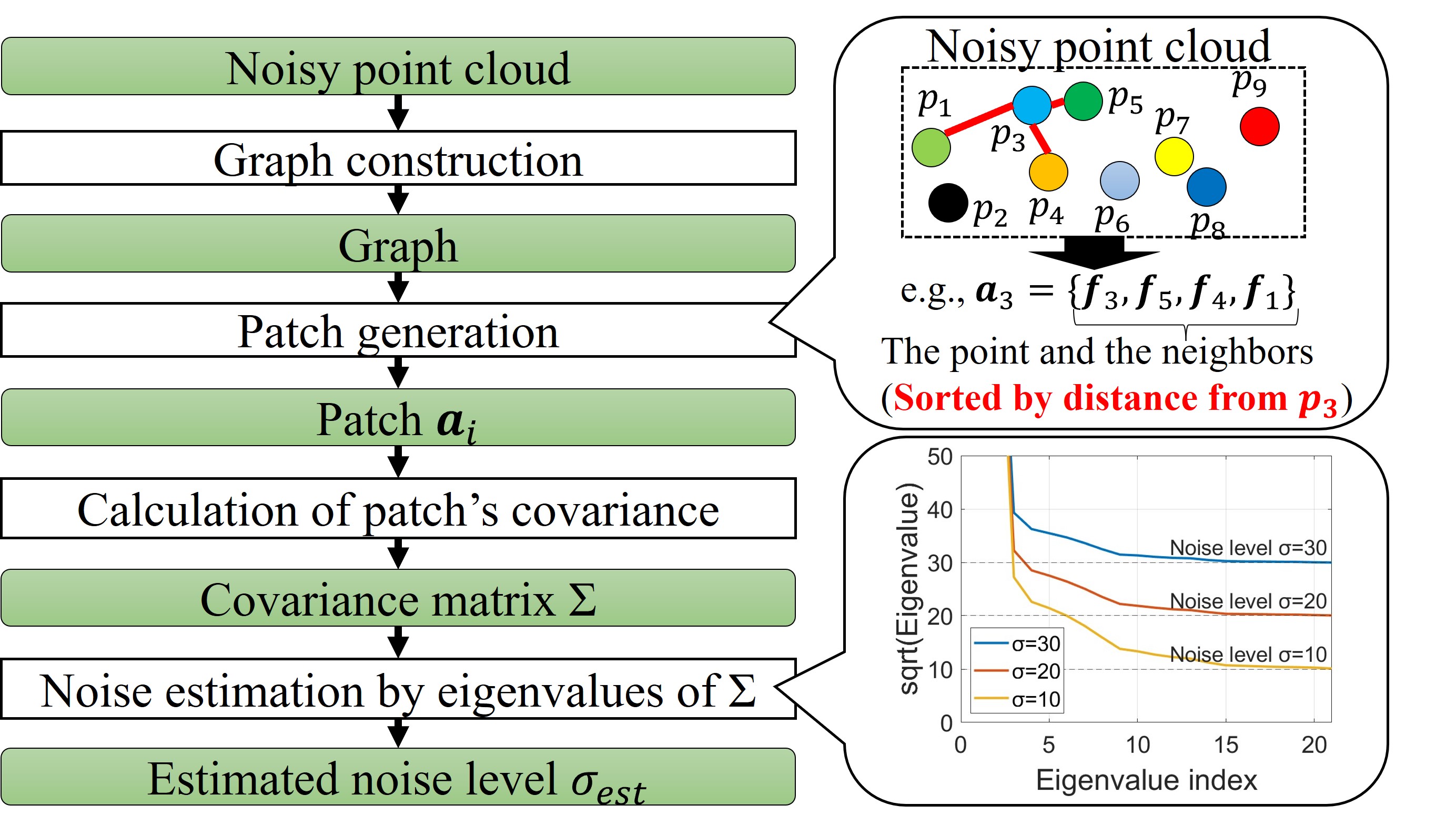}
   \caption{Noise estimation using graph-based patches (NE-GBP)}
\label{fig:ne-flow}
\end{figure}

\subsection{Noise estimation using graph-based patches (NE-GBP)}
\label{sec:proposed-NEGBP}

The problem of noise estimation has been widely studied in conventional 2-D image processing.
In particular, noise estimation methods using the covariance matrix of image patches achieve fast and accurate noise estimation results \cite{NEEigenvalue1, NEEigenvalue2, NEEigenvalue3}.
In these methods, after many small 2-D patches having the same size, e.g., blocks of $8 \times 8$ pixels, are taken from an image by a raster scan, the eigenvalues of the covariance matrix of patches are calculated.
Assuming that noise-free signals in small 2-D patches lie on a low-dimensional subspace, the noise variance can be estimated from the eigenvalues of principal components corresponding to higher frequencies that represent mostly noise.
However, the application of noise estimation methods for image processing to point clouds is not straightforward because of the irregular point cloud geometry. 
\gls{ne-gbp} utilizes the eigenvalues of the covariance matrix generated with small patches constructed with a graph. 
Fig.~\ref{fig:ne-flow} shows the flow of \gls{ne-gbp}.

First, a graph is constructed from a given noisy point cloud.
Although arbitrary graph construction methods can be applied, here, SLG described in Section \ref{sub-sec:proposed-scan-line-graph} is directly utilized.

Second, patches $\bm{A} = \{\bm{a}_i\}^N_{i=1} \in \mathbb{R}^{D \times N}$ are constructed 
from the attribute signal of $p_{i}$ and the neighbors from $p_{i}$ where $D$ is the number of signals in a patch.
As shown in Fig.~\ref{fig:ne-flow}, to better capture the statistical properties of different patches in the covariance matrix construction, the order of entries in vectors representing each patch should be consistent.  
Specifically, for any given patch vector, the entries (i.e., the attributes of the nodes) are ordered based on increasing distances of nodes within the patch. Thus, the first entry of each vector is the attribute of the query point, followed by the closest neighbor, and ending with the furthest neighbor. 
In sorted patches, the covariance matrix of patches $\bm{\Sigma} \in \mathbb{R}^{D 
 \times D}$ is calculated as 
$
\bm{\Sigma} = \frac{1}{N} \sum^{N}_{i=1} (\bm{a}_i - \bm{\mu}) (\bm{a}_i - \bm{\mu})^T
$
where $\bm{\mu} = \frac{1}{N} \sum^{N}_{i=1} \bm{a}_i$ is the mean of all patches.

Next, the noise level $\sigma_\mathrm{est}$ is estimated by using eigenvalues $\{\lambda_k\}^D_{k=1}$ ($\lambda_1 \geq \lambda_2 \geq \ldots \geq \lambda_D$) of the covariance matrix $\bm{\Sigma}$.
The original noise-free point cloud leads to patch covariance with eigenvalues that are close to zero for the subspace $S_\mathrm{H} = \{\lambda_k\}^D_{m+1}$. Thus, for a point cloud with noisy attributes, we propose to estimate the noise variance from the eigenvalues of $S_\mathrm{H}$. 
As derived theoretically in \cite{NEEigenvalue2}, $m$ should be chosen such that  
 $\tau = \frac{1}{D-m+1} \sum^{D}_{k=m+1} \lambda_{k} > 
 \text{median}(\{\lambda_{k}\}^{D}_{k=m+1})$.  
With this $m$ the noise level is calculated as $\sigma_\mathrm{est} = \sqrt{\tau}$.

\subsection{Filter selection and low-pass filter execution}
\label{sec:filter-selection}

In \gls{fgbd}, the low-pass filter is applied as follows:
\begin{table}[t]
\centering
\small
\caption{Processing time [s] of 3DPBS and FGBD in each section. The GC, NE, and LF indicate ``graph construction", ``noise estimation", and ``low-pass filter execution", respectively. These figures are average times per frame for all the frames and noise levels.}
\begin{tabular}{crrrrrr}
\hline
Method        & \multicolumn{2}{c}{3DPBS}                         & \multicolumn{2}{c}{FGBD-CPU}                      & \multicolumn{2}{c}{FGBD-GPU}                      \\ \hline
Dataset       & \multicolumn{1}{c}{8i} & \multicolumn{1}{c}{MVUB} & \multicolumn{1}{c}{8i} & \multicolumn{1}{c}{MVUB} & \multicolumn{1}{c}{8i} & \multicolumn{1}{c}{MVUB} \\ \hline
GC {[}s{]}    & 3.588                  & 1.769                    & 0.289                  & 0.095                    & 0.008                  & 0.005                    \\
NE {[}s{]}    & 17.342                 & 13.655                   & 0.912                  & 0.323                    & 0.032                  & 0.010                    \\
LF {[}s{]}    & 71.009                 & 24.270                   & 1.210                  & 0.395                    & 0.019                  & 0.011                    \\ \hline
Total {[}s{]} & 91.939                 & 39.693                   & 2.411                  & 0.812                    & 0.058                  & 0.026                    \\ \hline
\end{tabular}
\label{ref:tab-calc-time-CPUvsGPU}
\end{table}

\begin{table*}[t]
\centering
\small
\caption{The PSNRs [dB] of denoised point clouds and the processing time [s] calculated by the conventional and proposed methods.}
\begin{tabular}{ccrrrrrrrrr}
\hline
\multicolumn{1}{c}{\multirow{2}{*}{Dataset}} & \multicolumn{1}{c}{\multirow{2}{*}{$\sigma$}} & \multicolumn{1}{c}{\multirow{2}{*}{Noisy PSNR}} & \multicolumn{2}{c}{GLR \cite{DineshGLRGTV2019}}                             & \multicolumn{2}{c}{CD-SGW \cite{GAC}}                          & \multicolumn{2}{c}{3DPBS\cite{3DPBS}}                           & \multicolumn{2}{c}{FGBD-GPU}                        \\ \cline{4-11} 
\multicolumn{1}{c}{}                         & \multicolumn{1}{c}{}                   & \multicolumn{1}{c}{}                       & \multicolumn{1}{c}{PSNR} & \multicolumn{1}{c}{Time} & \multicolumn{1}{c}{PSNR} & \multicolumn{1}{c}{Time} & \multicolumn{1}{c}{PSNR} & \multicolumn{1}{c}{Time} & \multicolumn{1}{c}{PSNR} & \multicolumn{1}{c}{Time} \\ \hline
\multirow{3}{*}{8iVFBv2 \cite{8iVFBv2}}                     & 10                                     & 28.145                                     & 33.848                   & 6.618                    & 31.547                   & 22.305                   & \textbf{35.003}          & 97.283                   & 34.731                   & \textbf{0.021}           \\
                                             & 20                                     & 22.175                                     & 30.921                   & 7.199                    & 29.538                   & 21.620                   & \textbf{32.344}          & 88.505                   & 31.772                   & \textbf{0.024}           \\
                                             & 30                                     & 18.765                                     & 28.599                   & 6.847                    & 28.943                   & 21.387                   & \textbf{30.533}          & 90.031                   & 29.784                   & \textbf{0.024}           \\ \hline
\multirow{3}{*}{MVUB \cite{MVUB}}                        & 10                                     & 28.758                                     & 33.667                   & 2.417                    & 32.274                   & 8.261                    & 34.181                   & 36.247                   & \textbf{34.765}          & \textbf{0.010}           \\
                                             & 20                                     & 23.045                                     & 30.379                   & 2.832                    & 28.059                   & 8.314                    & 29.789                   & 34.248                   & \textbf{31.220}          & \textbf{0.011}           \\
                                             & 30                                     & 19.773                                     & 27.865                   & 3.510                    & 26.486                   & 8.435                    & 26.667                   & 48.587                   & \textbf{28.683}          & \textbf{0.012}           \\ \hline
\end{tabular}
\label{tab:ex-quantitative-results}
\end{table*}

\begin{equation}
\label{eq:LP-selfloop}
\boldsymbol{f}_\mathrm{out} = ({\mathbf{{D}}_\mathrm{g}}^{-1} {\mathbf{{W}}_\mathrm{g}})^q \boldsymbol{f}_\mathrm{in},
\end{equation} 
where $\boldsymbol{f}_\mathrm{in}$ and $\boldsymbol{f}_\mathrm{out}$ are input and denoised signals, respectively.
$q$ represents the number of iterations of the filtering.
Here, the adjacency matrix with self-loops $\mathbf{W}_\mathrm{g}$ is defined as $\mathbf{W}_\mathrm{g} = \mathbf{D} + \mathbf{W}$ where $\mathbf{D}$ is the degree matrix of $\mathbf{W}$, and $\mathbf{D}_\mathrm{g}$ is descried as $2 \mathbf{D}$.
According to the discussion in the previous study \cite{PCS2022}, 
the spectral interpretation of \eqref{eq:LP-selfloop} on the graph Fourier domain is described as $h_q (\lambda_i) = (1 - \frac{1}{2} \lambda_i)^q$.
In \gls{3dpbs}, the parameter $q$ is determined to satisfy the following condition based on $\sigma_\mathrm{est}$:
\begin{eqnarray}
\label{eq:afs}
\left| \sigma_\mathrm{est}^2 - \left(\frac{\Sigma_i^{N} \boldsymbol{y}[i]^2 - \Sigma_i^{N} \boldsymbol{x}_\mathrm{q}[i]^2}{N} \right) \right| < \epsilon,
\end{eqnarray}
where $\boldsymbol{y}[i]$ and $\boldsymbol{x}_\mathrm{q}[i]$ show observed noisy and denoised signals by the filter with parameter $q$, respectively.
$\epsilon$ is a small value defining a stopping criterion for the optimization process.
In \gls{3dpbs}, the selected filter realizes that the power of noise is equal to the power lost by the selected filter.
Since \gls{3dpbs} allows floating-point for $q$, a polynomial approximation is required to calculate \eqref{eq:LP-selfloop}.

Unlike \gls{3dpbs}, an integer value $q$ that minimizes \eqref{eq:afs} is selected in \gls{fgbd}. 
Thus, \eqref{eq:LP-selfloop} is simply calculated on the vertex domain which means that \eqref{eq:LP-selfloop} can be directly calculated by matrix operations.
While this simplification reduces processing time, the denoising accuracy is sometimes degraded.
To compensate for the quality of denoising, we propose an accurate filter selection method called \gls{fslr}.
The filter selection method shown by \eqref{eq:afs} is based on the assumption that high-frequency components contain only noise. 
However, there are some cases in which the optimum filter is not selected because high-frequency components include sharply changing edges in original signals.
The filter selection proposed in \cite{3DPBS} utilizes all the points for the calculation of \eqref{eq:afs}.
In contrast, some points in a point cloud are selected to calculate \eqref{eq:afs} in \gls{fslr}.
When \eqref{eq:afs} is calculated to decide the filter parameter $q$, we avoid utilizing the points that satisfy the following condition:
$(\sigma_R(i) + \sigma_G(i) + \sigma_B(i))/3    > 2\sigma_\mathrm{est}
$
where $\sigma_R(i)$, $\sigma_G(i)$, and $\sigma_B(i)$ indicate the standard deviations of red, green, and blue components in a patch $\bm{a}_i$, respectively.
To choose a more suitable low-pass filter, the point whose patch has a large color variance far exceeding the noise level is not used for filter selection. 

\section{EXPERIMENTS}
\label{sec:experiments}

\begin{table}[t]
\small
\centering
\caption{Processing time comparison of the graph construction methods with the 8iVFBv2 dataset. The figures indicate the average processing time per frame [s] in the graph construction process.}
\begin{tabular}{crrrr}
\hline
Method          & \multicolumn{1}{c}{BF-KNN \cite{BN-KNN}} & \multicolumn{1}{c}{kdtree \cite{KDtreeNN}} & \multicolumn{1}{c}{LBVH \cite{LBVH}} & \multicolumn{1}{c}{SLG} \\ \hline
8iVFBv2 & 339.069                    & 0.595                      & 0.528                    & \textbf{0.008}                   \\
MUVB   & 43.883                     & 0.226                      & 0.200                    & \textbf{0.005}                   \\ \hline
\end{tabular}
\label{tab:EX-graph-const}
\end{table}

\begin{table}[t]
\small
\centering
\caption{Comparison of the noise estimation error $E_\mathrm{ne} = |\sigma_\mathrm{est} - \sigma_\mathrm{act}|$ and processing time with the 8iVFBv2 dataset.}

\begin{tabular}{ccccc}
\hline
\multicolumn{2}{c}{Noise level} & $\sigma=$10           & $\sigma=$20           & $\sigma=$30           \\ \hline
\multirow{2}{*}{MEGW \cite{GAC}}                                                       & Error $E_\mathrm{ne}$& 1.894          & 1.854          & 1.051          \\
                                                                            & Time [s]  & 15.112         & 18.651         & 18.263         \\ \hline
\multirow{2}{*}{\begin{tabular}[c]{@{}c@{}}NE-GBP \\ w/o sort\end{tabular}} & Error $E_\mathrm{ne}$& 1.992          & 1.123          & 0.606          \\
                                                                            & Time [s] & \textbf{0.026} & \textbf{0.028} & \textbf{0.035} \\ \hline
\multirow{2}{*}{NE-GBP}                                                     & Error $E_\mathrm{ne}$ & \textbf{0.706} & \textbf{0.369} & \textbf{0.225} \\
                                                                            & Time [s]  & 0.028          & 0.030          & 0.037          \\ \hline
\end{tabular}
\label{tab:EX1-accuracy}
\end{table}

\subsection{Experimental conditions}
%
%
%
% [4.1.a Dataset]
\noindent \textbf{Datasets:} We evaluated all the frames in the 8iVFBv2 \cite{8iVFBv2} and \gls{mvub}~\cite{MVUB} datasets containing 1,200 and 1,202 frames.
Their color signals $\boldsymbol{f}_i$ were perturbed by additive Gaussian noise with standard deviation $\sigma=$ 10, 20, or 30.

%
%
%
% [4.1.b Evaluation metrics]
\noindent \textbf{Evaluation metrics:}
\gls{psnr} was measured in the same way as the previous study \cite{DineshGLRGTV2019}. 
To calculate the PSNR of the datasets \cite{8iVFBv2,MVUB}, we averaged each frame's PSNR.

\noindent \textbf{Computer specifications:} For measuring processing time, the computer which has Intel Core i9-9900K CPU @ 3.60GHz, NVIDIA RTX 2070, and 64GB RAM was utilized.

\subsection{Experimental results}
\label{sec:experiment}

\noindent \textbf{(1) Processing time of \gls{fgbd}:} 
First, we measured the processing time of each process of \gls{fgbd}, and compared it with that of \gls{3dpbs}~\cite{3DPBS}.
Since we implemented \gls{3dpbs} on a CPU, we prepared not only GPU implementation (FGBD-GPU) but also CPU implementation of \gls{fgbd} (FGBD-CPU) for a fair comparison.
Table \ref{ref:tab-calc-time-CPUvsGPU} shows that \gls{fgbd} can reduce the processing time from \gls{3dpbs} in all the processes.
Although GPU implementation is effective in accelerating the processing, even in the comparison between \gls{3dpbs} and \gls{fgbd}-CPU, \gls{fgbd}-CPU is much faster than 3DPBS.
Thus, the proposed algorithms were effective in accelerating denoising.

\begin{figure}[t]
\centering 
   %%%%%%%%%%%%%%%%%%%%%%%%%%%%%%%%%%%%%%%%%%%%%%%%%%%%%%%%%%%%%%%%%%%%%
   \subfigure[]{
   \includegraphics[width=.28\columnwidth]{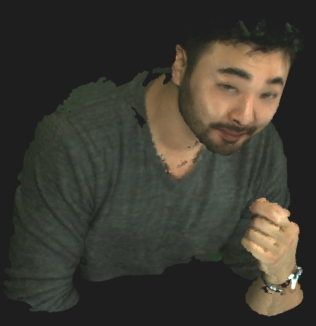}
   \label{fig:PC-GT}
  }
  \subfigure[]{
   \includegraphics[width=.28\columnwidth]{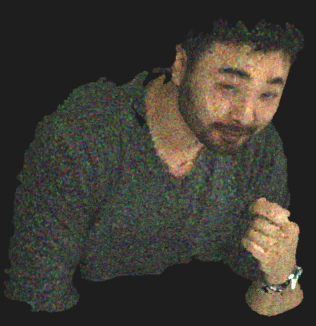}
   \label{fig:PC-NOISY}
  }
    \subfigure[]{
   \includegraphics[width=.28\columnwidth]{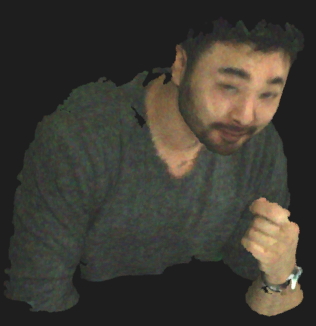}
   \label{fig:PC-GLR}
  }\\
  \subfigure[]{
   \includegraphics[width=.28\columnwidth]{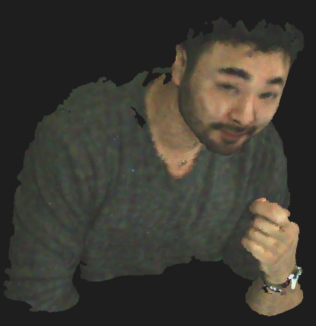}
   \label{fig:PC-SGWT}
  }
  \subfigure[]{
   \includegraphics[width=.28\columnwidth]{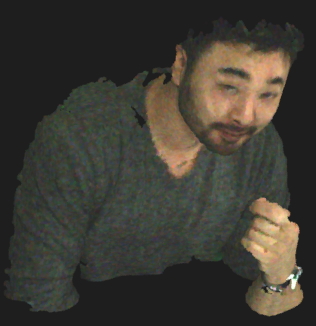}
   \label{fig:PC-BF}
  }
  \subfigure[]{
   \includegraphics[width=.28 \columnwidth]{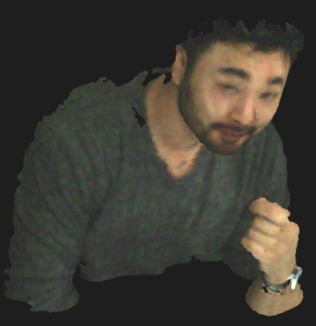}
   \label{fig:PC-PROPOSED}
  } 
  %%%%%%%%%%%%%%%%%%%%%%%%%%%%%%%%%%%%%%%%%%%%%%%%%%%%%%%%%%%%%%%%%%%%%
  \caption{The 1st frame of the "david" sequence in the MVUB with $\sigma = 30$: (a) ground truth, (b) noisy point cloud, denoising results by (c) GLR \cite{DineshGLRGTV2019}, (d) CD-SGW \cite{GAC}, (e) 3DPBS \cite{3DPBS}, and (f) FGBD.}
  \label{fig:PC}
\end{figure}

\noindent \textbf{(2) Comparison with conventional color denoising methods:} We compared \gls{fgbd} with the conventional methods: \gls{glr}~\cite{DineshGLRGTV2019}, \gls{cd-sgw}~\cite{GAC}, and \gls{3dpbs}~\cite{3DPBS}.
In this experiment, noise estimation of \gls{fgbd} was performed once every 10 frames. In the other frames, $q$ of the previous frame is utilized to accelerate the processing.
Table~\ref{tab:ex-quantitative-results} shows the quality of denoised point clouds and the processing time.
Besides, Fig.~\ref{fig:PC} shows denoised point clouds for the 1st frame of the ``david'' sequence in the MVUB dataset \cite{MVUB}.
According to Table~\ref{tab:ex-quantitative-results} and Fig.~\ref{fig:PC}, the denoising accuracy of \gls{fgbd} is comparable to those of the state-of-the-art denoising methods.
Besides, the processing time of \gls{fgbd} is faster than 30fps with both datasets.
%Note that the processing time for a frame where noise estimation is performed may be slower than 30 fps. However, this problem will be solved in practice by implementing noise estimation and low-pass filter execution in different threads.

\noindent \textbf{(3) Comparison with other graph construction methods: }
We compared the processing time of \gls{slg} and the conventional fast graph constructions, \gls{bf-knn}~\cite{BN-KNN}, kdtree~\cite{KDtreeNN} and \gls{lbvh}~\cite{LBVH}.
BF-KNN, \gls{lbvh}, and \gls{slg} are implemented on a GPU.
Table \ref{tab:EX-graph-const} shows that \gls{slg} outperformed the conventional methods in terms of computational complexity.

\noindent \textbf{(4) Evaluation of \gls{ne-gbp}:} 
We compared NE-GBP with the conventional noise level estimation method for point clouds called \gls{megw}~\cite{GAC} adopted in \gls{3dpbs}~\cite{3DPBS}.
Table~\ref{tab:EX1-accuracy} shows the comparison results of the noise estimation error $E_\mathrm{ne} = |\sigma_\mathrm{est} - \sigma_\mathrm{act}|$ and processing time.
Here, $\sigma_\mathrm{act}$ is the actual noise level.
As described in Section~\ref{sec:proposed-NEGBP}, the order of points in a patch is explicitly defined in \gls{ne-gbp}. To verify the importance of using a distance-based order, we compared our approach with a NE-GBP where the entries of the vector corresponding to each patch are not sorted (''NE-GBP w/o sort'' in Table~\ref{tab:EX1-accuracy}). 
According to Table~\ref{tab:EX1-accuracy}, 
the error and processing time of \gls{ne-gbp} are smaller than those of MEGW \cite{GAC}, and the order of points is important to acquire accurate noise levels.
In addition, when we used ''NE-GBP w/o sort'' for denoising, there was a 0.4 dB deterioration (average of all the point clouds and noise levels) compared with the PSNR of FGBD introduced in Table \ref{tab:ex-quantitative-results}.

\noindent \textbf{(5) Ablation study of \gls{fslr}:}
To verify the effect of \gls{fslr} introduced in Section~\ref{sec:filter-selection}, we compared the full \gls{fgbd} introduced in Table~\ref{tab:ex-quantitative-results} to \gls{fgbd} without \gls{fslr}.
Table~\ref{tab:FSLR} shows the comparison results for each dataset where we averaged the PSNR and processing time for all noise levels.
We observed an improvement of approximately 0.5dB for both datasets with a slight increase in processing time.

\begin{table}[t]
\small
\centering
\caption{The results of the abulation study of FSLR.}
\begin{tabular}{ccrcr}
\hline
\multirow{2}{*}{Dataset} & \multicolumn{2}{c}{FGBD (Full)}                                      & \multicolumn{2}{c}{FGBD w/o FSLR}                             \\ \cline{2-5} 
                         & PSNR {[}dB{]}              & \multicolumn{1}{c}{Time {[}s{]}} & PSNR {[}dB{]}              & \multicolumn{1}{c}{Time {[}s{]}} \\ \hline
8iVFBv2                  & \multicolumn{1}{r}{32.095} & 0.023                            & \multicolumn{1}{r}{31.691} & 0.021                            \\
MVUB                     & \multicolumn{1}{r}{31.556} & 0.011                            & \multicolumn{1}{r}{31.112} & 0.010                            \\ \hline
\end{tabular}
\label{tab:FSLR}
\end{table}
\section{CONCLUSION}
\label{sec:conclusion}

In this paper, we proposed a fast and accurate graph-based denoising method (FGBD).
Our three proposals, a scan line graph (SLG), noise estimation using graph-based patches (NE-GBP), and filter selection with limited region (FSLR), lead to real-time denoising with a single GPU while maintaining accuracy.
In the future, we will verify the effectiveness of \gls{fgbd} not only for color signals but also for coordinate signals to enlarge the scope of \gls{fgbd}.
%
%
%\vfill\pagebreak
\small
\bibliographystyle{IEEEbib}
\bibliography{ICASSP2024_ref}

\begin{thebibliography}{10}

\bibitem{telepresence}
Gerd Bruder, Frank Steinicke, and Andreas Nüchter,
\newblock ``Immersive point cloud virtual environments,''
\newblock in {\em 2014 IEEE Symposium on 3D User Interfaces (3DUI)}, 2014, pp. 161--162.

\bibitem{Holo3DTV}
Mostafa Agour and Thomas Kreis,
\newblock ``Experimental investigation of holographic {3D-TV} approach,''
\newblock in {\em 2009 3{DTV} Conference: The True Vision - Capture, Transmission and Display of 3{D} Video}, 2009, pp. 1--4.

\bibitem{2018PIXOR}
Bin Yang, Wenjie Luo, and Raquel Urtasun,
\newblock ``Pixor: Real-time 3{D} object detection from point clouds,''
\newblock in {\em 2018 IEEE/CVF Conference on Computer Vision and Pattern Recognition}, 2018, pp. 7652--7660.

\bibitem{ActionRecog2022}
Xing Li, Qian Huang, Tianjin Yang, and Qianhan Wu,
\newblock ``Hyperpointnet for point cloud sequence-based 3{D} human action recognition,''
\newblock in {\em 2022 IEEE International Conference on Multimedia and Expo (ICME)}, 2022, pp. 1--6.

\bibitem{PCC2018}
Sebastian Schwarz, Marius Preda, Vittorio Baroncini, Madhukar Budagavi, Pablo Cesar, Philip~A. Chou, Robert~A. Cohen, Maja Krivokuća, Sébastien Lasserre, Zhu Li, Joan Llach, Khaled Mammou, Rufael Mekuria, Ohji Nakagami, Ernestasia Siahaan, Ali Tabatabai, Alexis~M. Tourapis, and Vladyslav Zakharchenko,
\newblock ``Emerging {MPEG} standards for point cloud compression,''
\newblock {\em IEEE Journal on Emerging and Selected Topics in Circuits and Systems}, vol. 9, no. 1, pp. 133--148, 2019.

\bibitem{GPDNet}
Francesca Pistilli, Giulia Fracastoro, Diego Valsesia, and Enrico Magli,
\newblock ``Learning graph-convolutional representations for point cloud denoising,''
\newblock in {\em The European Conference on Computer Vision (ECCV)}, 2020, pp. 103--118.

\bibitem{DMRDenoise}
Shitong Luo and Wei Hu,
\newblock ``Differentiable manifold reconstruction for point cloud denoising,''
\newblock in {\em The 28th {ACM} International Conference on Multimedia}, 2020, pp. 1330--1338.

\bibitem{IBR}
Yann Schoenenberger, Johan Paratte, and Pierre Vandergheynst,
\newblock ``Graph-based denoising for time-varying point clouds,''
\newblock in {\em 2015 3DTV-Conference: The True Vision - Capture, Transmission and Display of 3D Video}, 2015, pp. 1--4.

\bibitem{GBPCD}
Xianz Gao, Wei Hu, and Zongming Guo,
\newblock ``Graph-based point cloud denoising,''
\newblock in {\em 2018 IEEE Fourth International Conference on Multimedia Big Data (BigMM)}, 2018, pp. 1--6.

\bibitem{GeomDenoising}
Dingkun Zhu, Honghua Chen, Weiming Wang, Haoran Xie, Gary Cheng, Mingqiang Wei, Jun Wang, and Fu~Lee Wang,
\newblock ``Nonlocal low-rank point cloud denoising for 3-{D} measurement surfaces,''
\newblock {\em IEEE Transactions on Instrumentation and Measurement}, vol. 71, pp. 1--14, 2022.

\bibitem{MS-GAT}
Xihua Sheng, Li~Li, Dong Liu, and Zhiwei Xiong,
\newblock ``Attribute artifacts removal for geometry-based point cloud compression,''
\newblock {\em IEEE Transactions on Image Processing}, vol. 31, pp. 3399--3413, 2022.

\bibitem{DineshGLRGTV2019}
Chinthaka Dinesh, Gene Cheung, and Ivan~V. Bajić,
\newblock ``3{D} point cloud color denoising using convex graph-signal smoothness priors,''
\newblock in {\em 2019 IEEE 21st International Workshop on Multimedia Signal Processing (MMSP)}, 2019, pp. 1--6.

\bibitem{GAC}
Muhammad~Abeer Irfan and Enrico Magli,
\newblock ``Joint geometry and color point cloud denoising based on graph wavelets,''
\newblock {\em IEEE Access}, vol. 9, pp. 21149--21166, 2021.

\bibitem{3DPBS}
Ryosuke Watanabe, Keisuke Nonaka, Eduardo Pavez, Tatsuya Kobayashi, and Antonio Ortega,
\newblock ``Graph-based point cloud color denoising with 3-dimensional patch-based similarity,''
\newblock in {\em ICASSP 2023 - 2023 IEEE International Conference on Acoustics, Speech and Signal Processing (ICASSP)}, 2023, pp. 1--5.

\bibitem{PCS2022}
Haoran Hong, Eduardo Pavez, Antonio Ortega, Ryosuke Watanabe, and Keisuke Nonaka,
\newblock ``Motion estimation and filtering for the inter prediction of dynamic point cloud compression,''
\newblock in {\em 2022 Picture Coding Symposium}, 2022.

\bibitem{BayesShrinkImage}
S.G. Chang, Bin Yu, and M.~Vetterli,
\newblock ``Adaptive wavelet thresholding for image denoising and compression,''
\newblock {\em IEEE Transactions on Image Processing}, vol. 9, no. 9, pp. 1532--1546, 2000.

\bibitem{8iVFBv2}
Eugene d'Eon, Bob Harrison, Taos Myers, and Philip~A. Chou,
\newblock ``8i voxelized full bodies - a voxelized point cloud dataset,''
\newblock in {\em ISO/IEC JTC1/SC29 Joint WG11/WG1 (MPEG/JPEG) input document WG11M40059/WG1M74006}, 2017.

\bibitem{MVUB}
C.~Loop, Q.~Cai, S.~O. Escolano, and P.~A. Chou,
\newblock ``Microsoft voxelized upper bodies - a voxelized point cloud dataset,''
\newblock {\em ISO/IEC JTC1/SC29 Joint WG11/WG1 (MPEG/JPEG) input document m38673/M72012}, 2016.

\bibitem{HighDimKNN1}
Xueyi Wang,
\newblock ``A fast exact k-nearest neighbors algorithm for high dimensional search using k-means clustering and triangle inequality,''
\newblock in {\em The 2011 International Joint Conference on Neural Networks}, 2011, pp. 1293--1299.

\bibitem{HighDimKNN2}
K~G Renga~Bashyam and Sathish Vadhiyar,
\newblock ``Fast scalable approximate nearest neighbor search for high-dimensional data,''
\newblock in {\em 2020 IEEE International Conference on Cluster Computing (CLUSTER)}, 2020, pp. 294--302.

\bibitem{BN-KNN}
Shengren Li and Nina Amenta,
\newblock ``Brute-force k-nearest neighbors search on the {GPU},''
\newblock in {\em Similarity Search and Applications}, Giuseppe Amato, Richard Connor, Fabrizio Falchi, and Claudio Gennaro, Eds., Cham, 2015, pp. 259--270, Springer International Publishing.

\bibitem{CUDA-KNN}
Vincent Garcia, Eric Debreuve, and Michel Barlaud,
\newblock ``Fast k nearest neighbor search using {GPU},''
\newblock in {\em 2008 IEEE Computer Society Conference on Computer Vision and Pattern Recognition Workshops}, 2008, pp. 1--6.

\bibitem{KDtreeNN}
Deyuan Qiu, Stefan May, and Andreas N{\"u}chter,
\newblock ``{GPU}-accelerated nearest neighbor search for 3{D} registration,''
\newblock in {\em Computer Vision Systems}, Mario Fritz, Bernt Schiele, and Justus~H. Piater, Eds., Berlin, Heidelberg, 2009, pp. 194--203, Springer Berlin Heidelberg.

\bibitem{LBVH}
J.~Jakob and M.~Guthe,
\newblock ``Optimizing {LBVH}-construction and hierarchy-traversal to accelerate k{NN} queries on point clouds using the {GPU},''
\newblock {\em Computer Graphics Forum}, vol. 40, no. 1, pp. 124--137, 2021.

\bibitem{radixsort}
Linh Ha, Jens Krueger, and Claudio~T. Silva,
\newblock ``{Fast Four-Way Parallel Radix Sorting on {GPU}s},''
\newblock {\em Computer Graphics Forum}, 2009.

\bibitem{NEEigenvalue1}
Xinhao Liu, Masayuki Tanaka, and Masatoshi Okutomi,
\newblock ``Single-image noise level estimation for blind denoising,''
\newblock {\em IEEE Transactions on Image Processing}, vol. 22, no. 12, pp. 5226--5237, 2013.

\bibitem{NEEigenvalue2}
Guangyong Chen, Fengyuan Zhu, and Pheng~Ann Heng,
\newblock ``An efficient statistical method for image noise level estimation,''
\newblock in {\em 2015 IEEE International Conference on Computer Vision (ICCV)}, 2015, pp. 477--485.

\bibitem{NEEigenvalue3}
Rui Chen, Fei Zhao, Changshui Yang, Yuan Li, and Tiejun Huang,
\newblock ``Robust estimation for image noise based on eigenvalue distributions of large sample covariance matrices,''
\newblock {\em Journal of Visual Communication and Image Representation}, vol. 63, pp. 102604, 2019.

\end{thebibliography}

\end{document}